\pdfoutput=1
\documentclass[11pt]{article}

\usepackage[]{acl}
\usepackage{times}
\usepackage{latexsym}
\usepackage{placeins}
\usepackage[T1]{fontenc}
\usepackage[utf8]{inputenc}

\usepackage{microtype}

\usepackage{booktabs}
\usepackage{graphicx}
\graphicspath{{graphs/}}

\newcommand{\gpts}{GPT2$_{tiny}$}
\newcommand{\gpt}{GPT2$_{small}$}
\usepackage{amssymb}

\usepackage[skip=4pt]{caption}
\usepackage{caption}
\usepackage{subcaption}
\usepackage{xcolor}
\newcommand{\cready}[1]{}

\usepackage{microtype}

\def\aclpaperid{1812} %  Enter the acl Paper ID here

\title{The Grammar-Learning Trajectories of Neural Language Models}

\author{ 
  Leshem Choshen\textsuperscript{\ensuremath\dagger}, Guy Hacohen\textsuperscript{\ensuremath\dagger}\textsuperscript{\ensuremath\ddagger}, Daphna Weinshall\textsuperscript{\ensuremath\dagger}, Omri Abend\textsuperscript{\ensuremath\dagger}\\
  Department of Computer Science\textsuperscript{\ensuremath\dagger}\\
  Department of Brain Sciences\textsuperscript{\ensuremath\ddagger}\\
  Hebrew University of Jerusalem\\
  {\tt\small \{first.last\}@mail.huji.ac.il}\\
  
}

\date{}

\begin{document}

\maketitle

\begin{abstract}

The learning trajectories of linguistic phenomena in humans provide insight into linguistic representation, beyond what can be gleaned from inspecting the behavior of an adult speaker. To apply a similar approach to analyze neural language models (NLM), it is first necessary to establish that different models are similar enough in the generalizations they make. 
In this paper, we show that NLMs with different initialization, architecture, and training data acquire linguistic phenomena in a similar order, despite their different end performance. These findings suggest that there is some  mutual inductive bias that underlies these models' learning of linguistic phenomena. Taking inspiration from psycholinguistics, we argue that studying this inductive bias is an opportunity to study the linguistic representation implicit in NLMs.

Leveraging these findings, we compare the relative performance on different phenomena at varying learning stages with simpler reference models. Results suggest that NLMs exhibit consistent ``developmental'' stages. Moreover, we find the learning trajectory to be approximately one-dimensional: given an NLM with a certain overall performance, it is possible to predict what linguistic generalizations it has already acquired.
Initial analysis of these stages presents phenomena clusters (notably morphological ones), whose performance progresses in unison, suggesting a potential link between the generalizations behind them. 

\end{abstract}

%%%%%%%%%%%%%%%%%%%%%%%%%%%%%%%%%%%%%%%%%%%%%%%%%%%%%%%%%%%%%%%
\section{Introduction}

%It is often argued that when children learn a language, they present similar developmental trajectories. Such similarities can be seen in . This fundamental phenomenon served as the premise of many linguistic theories, including  \citet{chomsky1957syntactic} and \citet{tomasello2000first}.

Children present remarkable consistency in their patterns of language acquisition. They often acquire linguistic phenomena in a similar order \citep{kuhl1992linguistic, ingram1989first}, and make similar generalizations and over-generalizations \citep{kuczaj1977acquisition, pinker1995language}. This consistency provides an important starting point for linguistic study. For example, arguments in favor of  single or dual system accounts of %account of 
morphological representation are often backed by computational models of children learning trajectories % presented by children with computational models 
%as in learning the past tense inflection in English
\citep[e.g.,][]{rumelhart1986on,pinker1988on,kirov-cotterell-2018-recurrent}. In this paper, we embrace this program for the study of computational language models, investigating learning trajectories. \footnote{Code is supplied in \url{https://github.com/borgr/ordert}} % We find NLMs' acquire language similarly, allowing analysis of what the trajectories imply about the linguistic structures learned by models. 

%Unlike the study of children, the study of NLM enjoys many advantages. In such a study, we can better understand the architecture, acquisition devices, modules and changes between different models. However, it is yet unclear whether LMs share a consistent learning dynamic. Even more obscure are the phases LMs go through during learning.

The representations that language models (LM) acquire have been studied extensively, including studying their learning dynamics to improve training (see \S\ref{sec:related_work}). However, very little work aimed at drawing connections between the training dynamics and the learned representations. %, presumably because there is limited similarity in representations between NLMs.
In this work we adopt a behavioral approach, thus revealing that NLMs share learning trajectories and generalize in similar ways during training. This implies that studying trajectories of NLMs is worthwhile, in the sense that results on one architecture or size are expected to be reproducible by others.

These findings call for a characterization of these trajectories, a new and promising territory for research. We take first steps to explore these directions, emphasizing their potential benefit to a better future understanding of what models learn.

Specifically, we train NLMs on next-word prediction, but evaluate and compare them by tracking their performance on grammar learning in English, using the BLIMP dataset (See \ref{sec:blimp}). 
BLIMP is a dataset that consists of 67K minimal pairs, where each pair includes a grammatically correct and a grammatically erroneous sentence. NLMs are tested for their ability to assign higher probability to the correct one.
See example in Table~\ref{tab:BLIMP}, and details of our experimental methodology in  \S\ref{sec:method}.

\begin{table}[t]
\small
\centering
\begin{tabular}{@{}p{2.1cm}p{2.4cm}p{2.4cm}@{}}
\toprule
Challenge & Correct & Erroneous \\ \midrule
Animate subject & \textbf{Galileo} had talked to Bell. & \textbf{This car} had talked to Bell. \\
% Animate subject & \textbf{Galileo} had talked to Stephanie. & \textbf{This car} had talked to Stephanie. \\
% \cmidrule{2-3}
% Determiner-noun agreement
% % \begin{tabular}[c]{@{}l@{}}Determiner Noun\\ agreement\end{tabular} 
% & Patricia bikes to those \textbf{malls}. & Patricia bikes to those \textbf{mall}. \\
\cmidrule{2-3}
Drop argument & The groups \textbf{buy}. & The groups \textbf{dislike}. \\ \bottomrule
% Drop argument & The committees \textbf{buy}. & The committees \textbf{dislike}. \\ \bottomrule
\end{tabular}
\caption{BLIMP minimal pairs examples.
% \vspace{-0.25cm}
}
\label{tab:BLIMP}
\end{table}

%We pose several questions: Do different LMs learn in the same order? If such order exist, what can we tell about it? are models with similar performance share the same linguistic generalizations?

%To address these questions, we consider a transfer learning framework. Regardless of the specific training task, we test our models on a set of semantic and morpho-syntactic challenges introduced as BLIMP \citep{Warstadt2019BLiMPTB}. BLIMP includes $67$ phenomena challenges, each containing $1000$ paired sentences of a specific linguistic phenomenon.
%Each pair contains sentences with minimal changes -- only the phenomenon in question is the source of difference between them (for example: ``she walk'' vs. ``she walks''). Such pairing allows cleaner evaluation.

We begin (\S\ref{sec:similar_process}) by establishing that NLMs learn grammatical phenomena in a consistent order. We evaluate NLMs at different time points along their training, showing that the performance on linguistic phenomena across initializations is highly correlated. We further find many similarities in the set of examples that they correctly classify.

Still, models of different architectures learn at a different pace, and hence cannot be directly compared at identical time points. In \S\ref{sec:full_models}, we overcome this %problem
by re-scaling the timeline. % based on the model's performance on the development set. 
We then show that despite architectural differences, NLMs present highly correlated performance trajectories. In \S\ref{sec:data}, we further demonstrate that even the choice of training data has minor influence on the results.
Finally, in \S\ref{sec:one_dimension} we show that the learning dynamics essentially follows a single dimension. Namely, where the average performance is similar, success on linguistic phenomena is also similar.
%Namely, at points where NLMs present a similar average performance, they fail and succeed on similar linguistic phenomena.
% %alternative a (models act)
% We conclude in \S\ref{sec:one_dimension} that there is essentially one dimension of learning. While off-the-shelf NLMs vary in their final performance, they share a constant learning order. 
% %alternative b
% We conclude in \S\ref{sec:one_dimension} that the learning dynamics follow a single dimension. Architectures and datasets, affect the pace of improvement along this dimension. Therefore, at points where NLMs present a similar average performance, they fail and succeed on similar linguistic phenomena.
% % We conclude in \S\ref{sec:one_dimension} that while off-the-shelf NLMs vary in their final performance, they share a constant learning order. 
% % %alternative b (order acts)
% % We conclude in \S\ref{sec:one_dimension} that while final performance reached varies between off-the-shelf NLMs, the order remains constant.
% % %original
% % We conclude in \S\ref{sec:one_dimension} that while off-the-shelf NLMs vary in their final performance, the order in which they learn remains constant.

We proceed by analyzing the early stages of learning in \S\ref{sec:phases}. We find that, at first, NLMs rely mostly on local cues and not on word order. They thus resemble bag-of-words models over a window of the preceding tokens. Later stages seem to drift further away from bag-of-words models toward $n$-gram models, and with time seem to be more sensitive to structural cues.
We also find evidence that some latent features that the model learns may not be related to linguistic phenomena.

Finally, in \S\ref{sec:types} we take the first steps in categorizing linguistic phenomena by their learning trajectories. We identify links between their representations by finding phenomena that progress in unison. For example, we find that morphological phenomena are mostly learned at similar stages. Of particular interest are cases where performance decreases with time, which may suggest either over-generalization or biases in the BLIMP challenges.

%%%%%%%%%%%%%%%%%%%%%%%%%%%%%%%%%%%%%%%%%%%%%%%%%%%%%%%%%%
\section{Experimental Setup}\label{sec:method}
\subsection{The  BLIMP Dataset}\label{sec:blimp}

We use BLIMP \citep{Warstadt2019BLiMPTB} to assess the  extent to which generalizations are made by the NLMs. 
%We hence use the BLIMP benchmark , created for that purpose. 
BLIMP includes 67 grammatical {\it challenges} categorized into 13 \textit{super-phenomena} (e.g., island-related or quantifiers) comprising of 4 broad \textit{fields} (e.g., Syntax, Semantics). Each challenge consists of 1K minimal pairs of sentences. A minimal pair contains a sentence and a near-duplicate distractor that incorporates an error on a particular linguistic phenomenon, i.e., only the phenomenon in question is changed between the sentences in a pair (see Table \ref{tab:BLIMP}). Each challenge includes pairs with the same linguistic phenomenon.

\subsection{Training}\label{sec:training_setup}

LM details: as training multiple GPT2 instances \citep{Radford2019LanguageMA} is computationally demanding, we train smaller NLMs. Following \citet{Turc2019WellReadSL}, we trained 1 instance of \gpt{} (width $768$, $12$ layers, $8$ attention heads) and 4 instances of \gpts{} (width $512$, $4$ layers, $4$ attention heads), with different random seeds.%\looseness=-1

Similarly, we train a small TransformerXL \citep{dai2019transformer}, $XL_{small}$ (width $512$, $4$ layers, $8$ attention heads) and a full-sized one (width $4096$, $18$ layers, $16$ attention heads). We stop the full model after 600K steps, while the perplexity remained high. We use it for comparison to the early stages of learning of TransformerXL. All models' hyperparameters can be found in App.~\S\ref{sec:settings}.
We also use the results of the fully trained GPT2, TransformerXL, LSTM and human performance reported in \citet{Warstadt2019BLiMPTB}. 

In \S\ref{sec:phases}, we compare NLMs with simpler models. To this end, we create two \gpts{} variations, denoted \textit{BOW} and \textit{Window-5}. BOW replicates \gpts{}, but relies only on bag of words. This is achieved by removing the positional weights, and replacing the attention weights with a simple average.\footnote{Supposedly, removing the positional embeddings would suffice. Empirically, it has little effect. Presumably, as embeddings only attend to previous positions, the network manages to represent positions by the difference between them. This is in line with the finding that GPT2's positional embeddings are not meaning-bearing \citep{wang2020position}.} Window-5 similarly ignores the positions, and additionally only attends to the last 5 words. Note that both are unidirectional LMs and consider only previously predicted words at each step. %Other scores used in \S\ref{sec:phases}, including other NLMs (LSTM, TransformerXL and GPT$_{large}$) and human performance, follow the previously reported values from \citet{Warstadt2019BLiMPTB}.

Unless explicitly stated otherwise (as in \S\ref{sec:data}), all models were trained on the WikiBooks dataset \citep[][]{Zhu2015AligningBA}, which contains the English Wikipedia ($2.1B$ words) and BookCorpus ($854M$ words). This dataset resembles BERT's training data \citep{Devlin2019BERT}, except that current Wikipedia is used. Additionally, we trained models on the following datasets: English openSubtitles \citep{Lison2016OpenSubtitles2016EL}, newsCrawl \citep{Barrault2019FindingsOT}, GigaWord \citep{napoles2012annotated}, and a sample of openWebText \citep[3B words;][]{Gokaslan2019OpenWeb} -- a replication of GPT2 dataset.

%For comparability, we trained all models on WikiBooks, except for the variants mentioned in \S\ref{sec:data}, that share architecture but not training data.

Throughout this paper, we report Pearson correlation. % are derived when 
Using Spearman correlation leads to qualitatively similar conclusions. When multiple models are correlated against each other, their average pair-wise correlation is reported.

\section{The Learning Order of NLMs}
\label{sec:similar_process}

In this section, we examine various aspects of NLMs, generally showing that their learning trajectories are similar.

We evaluate network similarity by adopting a behavioral approach. Accordingly, networks are viewed as functions, whose \textit{latent features} manifest themselves only by their influence on the network's behavior. Latent features are the unobserved causes of the measured behavior. Consequently, parameters, activation patterns and representations can be completely different among \emph{similar} models.
This is unlike the approaches employed by 
\citet{Williams2018DoLT,saphra2019understanding, Liu2021ProbingAT}, which analyze internal representations directly. 

To formalize the above notion, let $L_t$ denote a checkpoint, the language model $L$ at time $t$. Let $pv(L_t)$ denote its {\it performance vector} -- the accuracy obtained by $L$ on each BLIMP challenge $p$:
%\vspace{0.2cm}
 \begin{equation}
    pv(L_t) = [acc(L_t,p)]_{p \in BLIMP} \in \mathbb{R}^{67}
\end{equation}
Time $t$ is measured in training steps or perplexity.
The trajectory of the performance vector as a function of $t$ reflects $L$'s training dynamics. %The correlation between models is defined as the correlation between their average performance vectors.

Given this behavioral definition, we focus on comparing the relative strength of models. Similarity between models is thus measured as the correlation between their performance vectors. Hence, models are similar if they rank phenomena in the same way. 
%This is regardless of their average performance. 
On the other hand, models of the same average performance can be dissimilar: consider two models that agree on everything except nouns. One generates only feminine nouns and the other plural nouns.
%For example, consider 2 models that agree on all except gender. One model generates feminine nouns and the other masculine. 
The models' average performance is similar, but due to their biases, they are correct on different challenges. This dissimilarity suggests that the models rely on different latent features. 

\subsection{Consistent Order of Learning}

We begin by showing that models produced by different initializations learn the same phenomena, in the same order. In terms of our definitions above, this may imply that despite converging to different parameter values, the learned latent features and the generalization patterns made are similar.

\begin{figure}[tbhp]
\centering\includegraphics[width=\columnwidth]{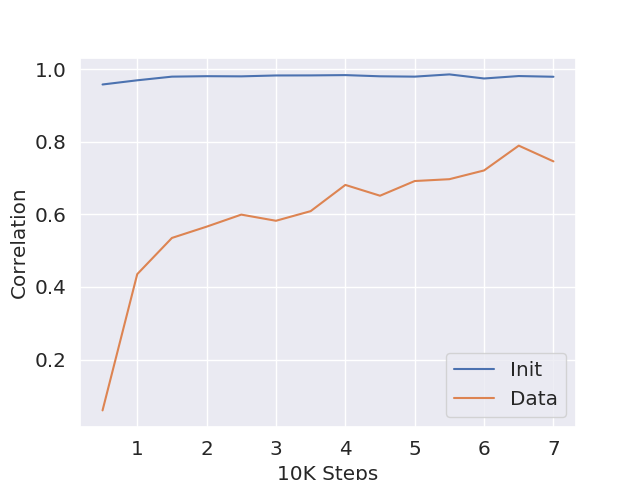}
    % \centering\includegraphics[width=0.8\columnwidth]{pearson_bothcorrelation_by_steps.png}
    % \captionsetup{aboveskip=0pt, belowskip=0pt}
    \caption{\textbf{High correlation} after warmup (5K steps). Correlation between the performance vectors (measured by steps) of \gpts{} models with different initialization (blue) or training data (orange)%. See discussion of the latter in  \S\ref{sec:full_models}
    .}
    \label{fig:correlation_by_step}
    % \vspace{-0.25cm}
\end{figure}

In order to examine the hypothesis empirically, we compute 
the correlation between 4 random initializations %. We plot their correlation as a function of the number of steps 
(Fig.~\ref{fig:correlation_by_step}). Results confirm the hypothesis, the correlation between \gpts{} instances is extremely high. It is already high after 10K steps, and remains high throughout training. We note that the correlation at step 0 is 0 (not shown), and that after 10K warm-up steps the network's ability as a LM is still poor. For example, perplexity is 10.9 after 10K steps and 6.7 after 70K steps. 

%%%%%%%%%%%%%%%%%%%%%%%%%%%%%%%%%%%%%%%%%%%
\subsection{Effects of Architecture}

\begin{figure}[htbp]
    \centering\includegraphics[width=\columnwidth]{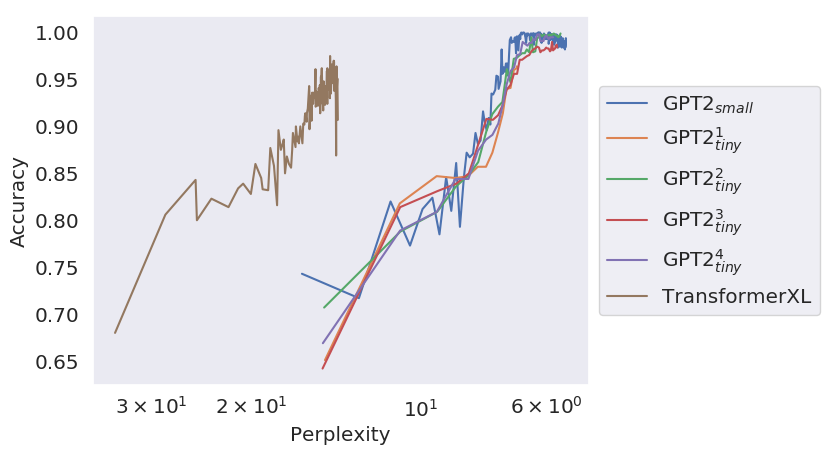}
    \caption{\textbf{Similar Accuracy} despite different initializations and sizes of the \gpt{} models. TransformerXL perplexity is not computed on the same vocabulary, but still shows a (rescaled) similar trend.
    The graph depicts trajectories on an example phenomenon (``existential there'').
    y-axis is the accuracy during training and x-axis is the model's perplexity.}
    \label{fig:challenge_example}
    % \vspace{-0.25cm}
\end{figure}

\noindent
Next, we show that different architectures also present similar trajectories.
As the learning pace is not comparable across models, computing correlation in fixed and identical intervals is not informative. Instead, we choose $t$ to be the perplexity on the development set, comparing models at the same performance level. TransformerXL is not directly comparable as perplexity requires the vocabulary to be the same.

Following this paradigm, we see that \gpt{} and \gpts{} are highly correlated (>$0.9$), presenting similar learning order throughout training.
Observing the trajectories per challenge qualitatively, we see that they align very well (cf. Fig.~\ref{fig:challenge_example} and App. \S\ref{per_challenge}, \S\ref{sec:during_train}). TransformerXL also seems to share the general tendencies of the GPT2 architectures. 

Interestingly, we see that models behave similarly not only in terms of relative performance, but also at the example level (binary decision per minimal pair). We find that \gpt{} and \gpts{} have an average agreement of $\kappa=0.83$ \citep{Fleiss1969LargeSS}. This implies strong consistency in the order of learning of different examples also within phenomena. Henceforth, we focus on the phenomena-level as it is more interpretable, lending itself more easily to characterization. We discuss per-example similarity further in App.~\S\ref{sec:example_corr}.

\subsection{Comparison to Off-the-shelf Models}\label{sec:full_models}

\begin{figure*}
\begin{subfigure}{\columnwidth}
    \centering\includegraphics[width=\columnwidth]{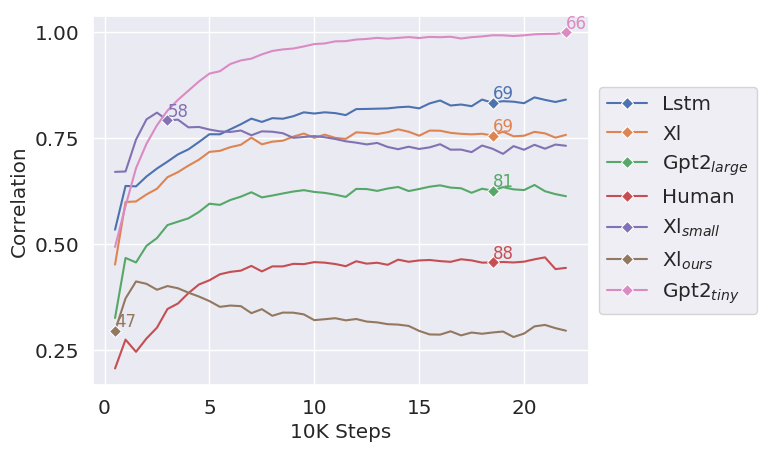}
    \caption{Off-the-shelf and human\label{fig:best_metrics}
    % \vspace{-0.1cm}
    }
\end{subfigure}
\begin{subfigure}{\columnwidth}
    \centering\includegraphics[width=\columnwidth]{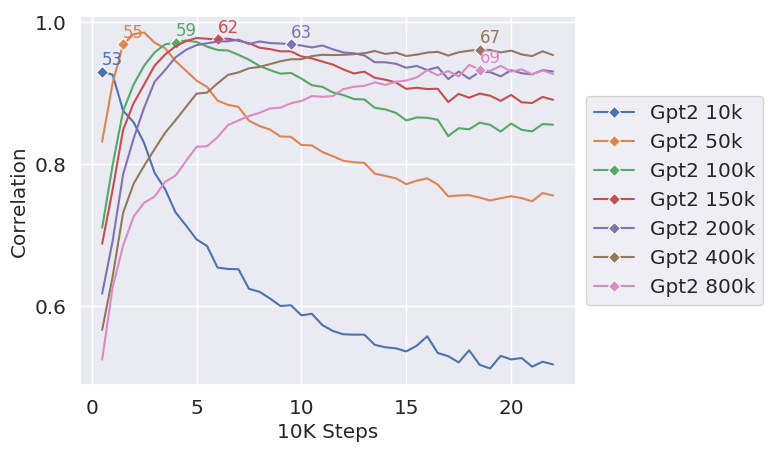}
    \caption{\gpt{} checkpoint after $X$ steps \label{fig:gpt2steps}
    % \vspace{-0.1cm}
    }
\end{subfigure}

    \caption{Reference models \textbf{correlate} the most with \gpts{} \textbf{when} they have the most \textbf{similar performance} (or near it). 
    Correlation during \gpts{} training compared to off-the-shelf LMs and human performance (left) or to mid-training \gpt{} checkpoints  (right). Curves correspond to fixed performance vectors. Where the X-axis follows the training trajectory of gpt, each line represents similarity to a different checkpoint, either of different fully trained models (left) or to checkpoint during the training of a larger model (right).
    Numbers are the average performance of the checkpoint, and are placed over the step where this average performance is the most similar to that of \gpts{}. The best score of \gpts{} is 67. 
    % \vspace{-0.25cm}
    }
    \label{fig:my_label}
\end{figure*}

So far, we have observed the common trajectories presented by NLMs that are trained in parallel. We proceed to compare trajectories of one model to other models' performance vectors at a single point of interest in their learning, i.e. a checkpoint's performance vector. This allows us to analyze how similarities evolve, rather than whether two trajectories are synced. We compare fully trained off-the-shelf NLMs with the trajectory of %our in-house trained
\gpts{} (Fig. \ref{fig:best_metrics}) and \gpt{} (App. \S\ref{sec:other_models}).%\looseness=-1 %As off-the-shelf models we take LSTM, TransformerXL and GPT$_{large}$ public performance vectors, and compare them to \gpts{} (Fig. \ref{fig:best_metrics}) and \gpt{} (App. \S\ref{sec:other_models}).

%to LSTM, TransformerXL and off-the-shelf models. We compare throughout their training, but only to models'. 

The observed similarity to off-the-shelf models is high (0.6-0.8), implying that NLMs in general share tendencies and biases. Moreover, similarity increases until the point of same performance and then (when relevant) decreases. This suggests that the small NLM approaches off-the-shelf tendencies as it improves and stops somewhere on the same trajectory of generalizations (cf. \S\ref{sec:one_dimension}). Furthermore, we find considerable correlation with the performance levels of humans on the different challenges, but still, all NLMs correlate better with our model than humans correlate with it.

% \begin{figure}[tbhp]
%     \centering\includegraphics[width=.8\columnwidth]{pearson_blimpmetrics.png}
%     % \captionsetup{aboveskip=0pt, belowskip=0pt}
%     \caption{Correlation between the difficulty predicted by BLIMP models and the difficulties for the model for each phenomena in each training step.}
%     \label{fig:blimp-metrics}
% \end{figure}

% \begin{figure}[tbhp]
%     \centering\includegraphics[width=\columnwidth]{pearson_bestmetrics.png}
%     \caption{Correlation during training of \gpts{} compared to a fixed performance vector, derived either from other off-the-shelf LMs, or human performance.
%     Curves correspond to different fixed performance vectors.
%     Numbers on the curves are the average performance of the fixed vectors, and are placed over the step where this average performance is the most similar to the average performance of \gpts{}. The correlation with \gpts{}'s own end-state is shown over \gpts{}. Results show that reference models correlate the most with \gpts{} when they have the most similar performance (or near it).  The best score of \gpts{} is 67. {\color{red} increase font size}}
%     \label{fig:best_metrics}
% \end{figure}

These results present a curious order imposed on the NLMs. Both \gpts{} and \gpt{} (App. \S\ref{sec:other_models}) are more similar to the LSTM model than to TransformerXL, and even less similar to GPT2$_{large}$. Interestingly, our models are more similar to an RNN and a model with a different architecture, than to a larger model with the same architecture. 
Thus, it seems that the architecture type cannot explain the similarities in the relative order. We further examine this issue in the next section.

%The next sections will examine whether the training data can explain Our models are more similar to models with more similar average performance. 

%%%%%%%%%%%%%%%%%%%%%%%%%%%%%%%%%%%%%%%%%%%%%%%%%%%%%%%%%%%
\subsection{Effect of Training Data} \label{sec:data}

This section examines the possibility that 
the similarities reported in Fig.~\ref{fig:best_metrics} can simply be explained by the similarity in the NLM's training data.
More specifically, since the ranking by model similarity reported above fits the similarity between the training sets that the models were trained on, we view it as a potential confound and attempt to control for it.
Our training data (WikiBooks) consists mostly of Wikipedia and so do the LSTM's and TransformerXL's training sets, which are trained on earlier versions of Wikipedia and WikiMatrix \citep{schwenk2019wikimatrix} respectively. GPT2, on the other hand, is trained on openWebText, which consists of scraped web pages.
%, filtered in order to avoid auto-generated content.

To tease apart the effect of training data, we trained 3 additional \gpts{} instances over the openWebText, openSubtitles and newsCrawl datasets. Results (Fig.~\ref{fig:correlation_by_step}) show that the dataset has more effect on the correlation than initialization. Hence, the choice of training data does affect the learning trajectory, but its effect decreases with training (correlation gets higher with more training steps). We also recompute the correlations from \S\ref{sec:full_models} after training \gpts{} on the same data as GPT2$_{large}$ (App. \S\ref{sec:other_data}), and find that the relative order between the NLMs remains the same, with GPT2$_{large}$ being the least similar.

We conclude that while the training data affects the learned generalizations, it only very partially explains the observed similarities between NLMs. 
%In other words, which latent features matter to the model depends mainly on something other than data.

\subsection{One Dimension of Learning}\label{sec:one_dimension}

Based on the findings of the previous sub-sections, we hypothesize that current NLMs all learn in a similar order, where the effect of training data and architecture is secondary. In other words, training time, size and efficiency may affect what a model has learned, but not its learning order. This implies that stronger models may improve performance, but still follow a similar learning trajectory. If this hypothesis is correct, models should be most similar to models with the same performance; similarity should drop as the gap in performance widens. 

% \begin{figure}[tbhp]
%     \centering\includegraphics[width=\columnwidth]{pearson_gpt2stepsmetrics.png}
%     \caption{Correlation during training of \gpts{} compared to steps in training of \gpt{}. Correlation is over BLIMP challenges. Numbers indicate the overall average of the reference models over BLIMP, and are computed over the step with most similar accuracy on \gpts{}. The best score of \gpts{} is 67.}
%     \label{fig:gpt2steps}
% \end{figure}

Controlled comparison supports this hypothesis.
Fig.~\ref{fig:gpt2steps} presents the correlation of \gpts{} training trajectory with several static checkpoints taken during \gpt{} training.
We observe that at the point in which the average performance of \gpts{} is closest to that of the checkpoint, the correlation peaks, and then decreases again as \gpts{} surpasses the checkpoint in average performance. So overall correlation peaks when average performance is most similar.
%Moreover, models that outperform \gpts{} are less similar to it. 
Note that despite the different network sizes and convergence rates, the correlation's maximal value is very high (higher than 0.9).

Further experiments show similar trends.
Fig. \ref{fig:best_metrics} presents a similar investigation, albeit with more varied architectures and training datasets. 
Here too the maximum correlation is obtained around the point of most similar performance.
%This may explain why LSTM, the smallest and least effective model, is more similar to \gpts{} than the larger and better GPT2$_{large}$.

\subsection{Comparison to 5-gram}\label{sec:ngrams}

NLMs are most similar to other NLMs with the same performance. However, when compared to non-neural LMs, this is no longer the case.

\begin{figure}[tbhp]
    \centering\includegraphics[width=\columnwidth]{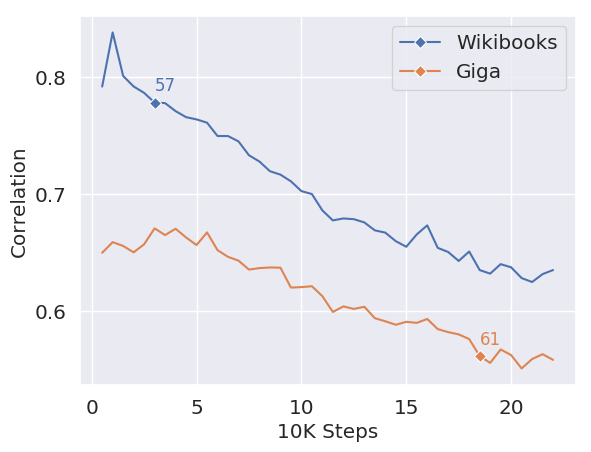}
    % \centering\includegraphics[width=0.8\columnwidth]{pearson_5grammetrics.png}
    \caption{Correlation during training of \gpts{} compared to a \textbf{5-gram} model trained on the same data (WikiBooks) and on GigaWord. On each curve, we mark the point at which the accuracy is most similar to \gpts{}, and additionally indicate the corresponding overall average accuracy of the reference models.
    %\vspace{-0.25cm}
    }
    \label{fig:5gram}
\end{figure}

More specifically, we compare \gpts{} to two 5-gram LMs trained on the same dataset as the NLMs (WikiBooks) and another (GigaWord) dataset. Results are shown in Fig. \ref{fig:5gram}, which is qualitatively different from Fig. \ref{fig:best_metrics}. Here, similarity in performance implies neither high correlation, nor the point of highest similarity. This serves both as a sanity check to our methodology, and as a reminder of model biases: In general, models may have different biases and tendencies, regardless of overall performance. In our case, it seems that NLMs share biases between them that are not necessarily shared with other LMs.

While not the main purpose of the analysis, our comparison reveals other noteworthy trends. For example, 5-gram LMs trained on different corpora have different correlations to the \gpts{} trajectory. This is further discussed in App. \S\ref{sec:5grams}.

\subsection{Discussion}

We find that the order of learning is surprisingly stable across architectures, model sizes and training sets. Therefore, given a new NLM, the order in which it will learn linguistic phenomena can be predicted by another model that achieves a similar average accuracy. When considering non-neural LMs, this observation does not always hold: inherently different architectures (such as 5-grams) have very different trajectories. Hence, future models with very different induced biases may present different orders.

% (App. \S\ref{sec:other_data})

%%%%%%%%%%%%%%%%%%%%%%%%%%%%%%%%%%%%%%%%%%%%%%%%%%%%%%%%%%%%%%%%%%%%
\section{Phases of Learning}\label{sec:phases}

Having established that different NLMs learn in a consistent order, we investigate the emerging learning trajectory by comparing it with simpler reference models. Our goal is to identify distinct learning phases that characterize NLM's training.

\paragraph{Setup.}
We compare \gpts{} to fully trained LMs (same as \S\ref{sec:full_models}), as well as to a variety of metrics. For each metric $m$ we compute the average score over each example for each of the 67 sets $\mathbb{E}_{p_i\in p}\left[m\left(p_i\right)\right]\in \mathbb{R}^{67}$. The results are replicated with \gpt{} and TransformerXL and lead to similar conclusions (see App.~\S\ref{sec:other_models}).

\paragraph{Sentence-level Metrics.}
First, we consider two sentence-level metrics: sentence length (in tokens) and syntactic depth. Assuming a sentence parse tree, the depth is the longest path from a word to the root. Sentence length is often considered to be a source of challenge for infants \citep{brown1973first} and networks \citep{Neishi2019OnTR}, regardless of the sentence's complexity. Syntactic depth \citep{yngve1960model} is a measure used to assess how cognitively complex a sentence is. We leave the question of which measure of linguistic complexity \citep{szmrecsanyi2004operationalizing} correlates best with the trajectory exhibited by NLMs to future work.

\begin{figure}[tbhp]
    \centering\includegraphics[width=\columnwidth]{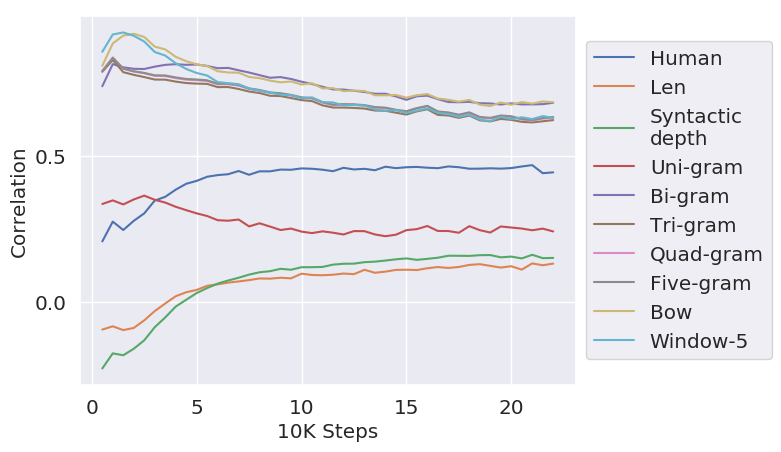}
    % \captionsetup{aboveskip=0pt, belowskip=0pt}
    \caption{Correlation between the performance vectors of different \textbf{metrics} and \textbf{models} against the vector of \gpts{} at different stages of learning.}
    % \vspace{-0.25cm}}
    \label{fig:metrics}
\end{figure}

%Figure \ref{fig:metrics} presents the correlation of \gpts{} with the different models and metrics. 
Our results (Fig.~\ref{fig:metrics}) show that neither sentence-level metric (length and syntactic depth) can predict well what is difficult for the model.
This is not surprising, as both measures only capture sentence complexity at a general level, and are not directly related to the linguistic phenomenon that is being tested. We do see that the syntactic depth starts off as a worse predictor of the NLM performance and ends as a better one. We provide a different perspective on this initial learning phase, before and after that switch, later in this section.

% \oa{I would remove the following paragraph; it's very speculative, and we have much stronger results in the other parts.}
% The results of this comparison suggest that structure plays a role in the network's prediction. Despite the similarity of the two metrics in BLIMP,\footnote{BLIMP contains little redundancy. Hence, there aren't many adjectives conjunctions and other words that make the sentence longer but does not affect the syntactic tree length.} they do differ in their predictive ability. Thus the length seems more relevant at the beginning, while depth is more relevant later on. As depth is an aggregation of syntactic features, the fact that it better predicts the difficulty of phenomena implies that the network relies on related features in its prediction.

%\paragraph{Comparison with .}
Next, we compare the performance vector with task difficulty for humans, as reported in the original BLIMP paper. We observe that correlation is fairly high after a sufficient number of steps. In fact, the network becomes more similar to humans as it improves: at the beginning, the network relies on different features than humans, but with time more of the hurdles are shared. However, correlation saturates at a mid-range correlation of under  0.5. This suggests that the network (partially) relies on features that are not used by human annotators. These may be valid generalizations not tested by BLIMP, or erroneous ones that are still beneficial to reduce the score on the task it was trained on \citep[cf.][]{mccoy2019right}. We revisit this issue in \S\ref{sec:types}.

\paragraph{Comparison with Limited Context and Locality.}
Our methodology opens the door to examine other potential biases of LMs. We now do so, starting with context and locality.

We consider models that take into account different scopes of context: unigram, and 2-5 gram LMs that can exploit the order of preceding words. We argue that the correlation between NLMs and $n$-gram LMs may indicate that features based on limited context are also employed by NLMs. 
%challenges they are better at are challenges that can be solved by this amount of context (See also \S\ref{sec:types}).

Surprisingly, the unigram model, which doesn't use context, perfectly classifies 7 phenomena, achieves 98.1\% accuracy on 1, and completely fails (0\% accuracy) on 8. This suggests that high accuracy on some syntactic and semantic challenges (as defined by BLIMP) can be achieved by simple heuristics. Note, however, that the NLMs we test are not trained towards any specific phenomena and are not fine-tuned in any way. Hence, NLMs can only attain heuristics or biases (generalization errors) which are beneficial in general, not ones specific to our test challenges. 

While NLMs initially present a strong correlation with the unigram model, this correlation quickly drops (see Fig.~\ref{fig:metrics}). 
From the outset, \gpts{} succeeds on 6 of the 8 phenomena that are classified well by unigrams, and 4 of the 8 that the unigram model utterly fails on. Interestingly, for 3 of the other phenomena on which the unigram failed, \gpts{} initially achieves 0\% accuracy (chance level is 50\%), but its accuracy does climb during training (e.g., see App. \S\ref{per_challenge}). We conclude that, as expected, the NLM acquires a bias towards predicting frequent words early in training, but that this bias is weighed in against other (contextual) considerations later on in training.

\begin{figure}[tbhp]
\centering\includegraphics[width=\columnwidth]{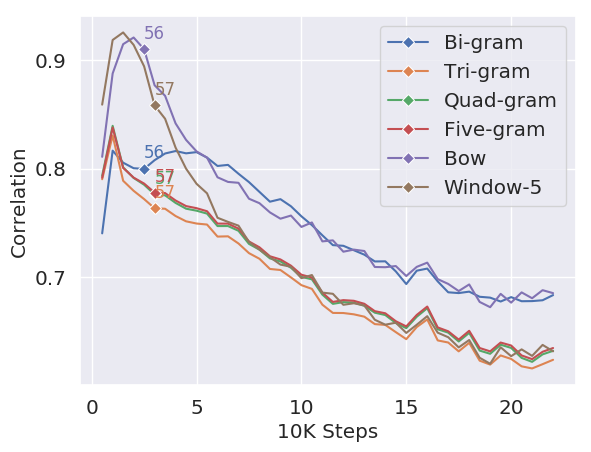}
    % \centering\includegraphics[width=0.8\columnwidth]{pearson_modelsmetrics.png}
    % \captionsetup{aboveskip=0pt, belowskip=0pt}
    \caption{Correlation between the performance vectors of \gpts{} throughout learning with simple LMs. The figure focuses on LMs found also on Fig. \ref{fig:metrics}.
    % \vspace{-0.25cm}
    }
    \label{fig:model_metrics}
\end{figure}

%%%%%% |||| %%%%%
Comparing different scopes of context, our results (Fig. \ref{fig:model_metrics} and App. \S\ref{sec:other_models}) show that throughout training, the network presents high correlation with $n$-gram models. %Among the models, similarity to 3-gram, 4-gram and 5-gram models is initially higher, but at a certain point, the network becomes most similar to the bi-gram model.
From a certain point onward, the network becomes more similar to the bi-gram model than to the other $n$-gram LMs. %Correlation with $n$-gram models (for $n>1$) proved to be among the strongest predictors.
%is also a much better predictor than what is challenging for humans. 
We also note that similarity peaks early on, but with time the correlation decreases. This may suggest that initially, the NLMs acquire grammatical behavior that resembles a Markov model, or even a bi-gram model. Only later does the network rely more on global features. This is in line with our earlier findings, which show an increasing correlation with syntactic depth as compared to sentence length.

At the very beginning, NLMs often generate one word repetitions \citep[e.g., "the"][]{fu2020theoretical}.
This seems to be at odds with our finding that grammar learning already begins at this early stage. However, while frequency may dictate the most probable predictions, comparing two options that differ only slightly may prove to depend more on context, as our results indicate.

\paragraph{Limited Context and Word Order.}
By comparing NLMs to $n$-grams, we examined the effect of context within a fixed window size. Now we examine the effect of word order, within a window and in general. To this end, we create two ablated \gpts{} models. BOW is agnostic to the order between preceding tokens, while Window-5 is similar but relies only on 5 tokens (details in \S\ref{sec:method}). 

Our results suggest that initially, the identity of the preceding words is more important than their order. Both BOW and Window-5 better correlate with our NLM than the $n$-gram models. Later on, this trend reverses and the $n$-grams, that do exploit word order, become better correlated. Furthermore, the correlation with Window-5 is significantly smaller than with BOW at later stages of learning, suggesting that the network gradually learns to rely on more context \citep[cf.][]{saphra2019understanding}. 
%We note that although these are neural models, their behavior differs from other NLMs due to the inherent bias imposed by their architectures.
%(for behaviour of other models \S\ref{sec:full_models} and \S\ref{sec:data}).

%%%%%%%%%%%%%%%%%%%%%%%%%%%%%%%%%%%%%%%%%%%%%%%%%%%%%%%%%%%%%%%%%%%%

\begin{figure*}[tbh]
    \begin{subfigure}{.245\textwidth}
      \centering
      \includegraphics[width=1\linewidth]{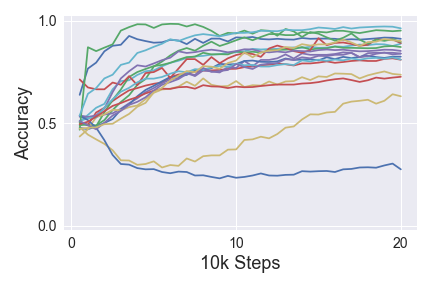}
      \caption{Morphology}
      \label{fig:field_morphology}
    \end{subfigure}
    \begin{subfigure}{.245\textwidth}
      \centering
      \includegraphics[width=1\linewidth]{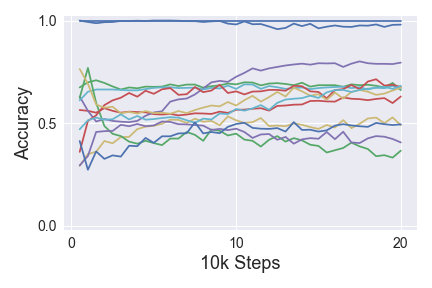}
      \caption{Syntax-Semantics}
      \label{fig:field_syntax_semantics}
    \end{subfigure}
    \begin{subfigure}{.245\textwidth}
      \centering
      \includegraphics[width=1\linewidth]{spectral_clusters/field/semantics.png}
      \caption{Semantics}
      \label{fig:field_semantics} 
    \end{subfigure}
    \begin{subfigure}{.245\textwidth}
      \centering
      \includegraphics[width=1\linewidth]{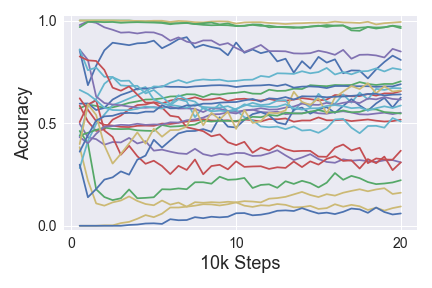}
      \caption{Syntax}
      \label{fig:field_syntax}
    \end{subfigure}

    \caption{Morphology and Syntax-Semantics (left) characterize NLM learning well, while semantics and syntactic phenomena show little similarity (between lines). Learning curves of \gpts{} per challenge (lines), clustered according to different fields (graphs) and colored by super-phenomena. 
    % \vspace{-0.25cm}
    }
    \label{fig:field_clusters}
\end{figure*}

%Further clustering "semantics" and "syntax" fields according to super-phenomena do not solve this problem. While some super-phenomena depict exhibits similar behaviors, most of them lack a typical learning curve, suggesting other categorization schemes may better characterize the learning (see App.~\S\ref{app_sec:all_clusters})
\begin{figure*}[th]
\begin{subfigure}{.245\textwidth}
      \centering
      \includegraphics[width=1\linewidth]{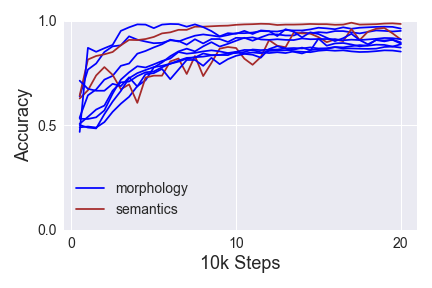}
      \caption{}
      \label{fig:cluster_start_random_and_improve}
    \end{subfigure}
    \begin{subfigure}{.245\textwidth}
      \centering
      \includegraphics[width=1\linewidth]{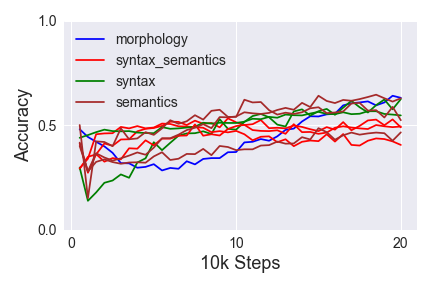}
      \caption{}
      \label{fig:cluster_near_change_level}
    \end{subfigure}
    \begin{subfigure}{.245\textwidth}
      \centering
      \includegraphics[width=1\linewidth]{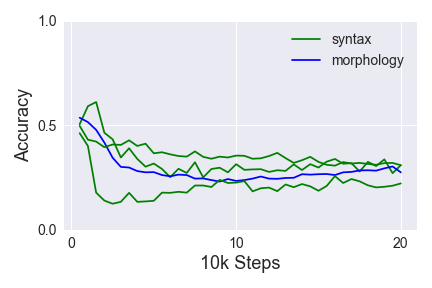}
      \caption{}
      \label{fig:cluster_deteeriorate}
    \end{subfigure}
    \begin{subfigure}{.245\textwidth}
      \centering
      \includegraphics[width=1\linewidth]{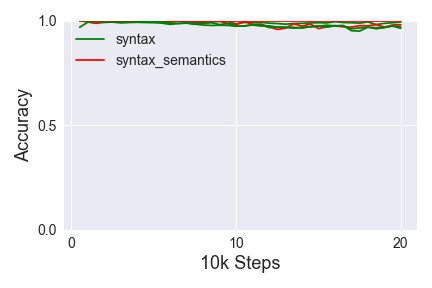}
      \caption{}
      \label{fig:cluster_perfect_accuracy}
    \end{subfigure}

    \caption{Some phenomena are learned, others (\subref{fig:cluster_deteeriorate}) deteriorate, implying the network (that learns language modelling, not phenomena) learns orthogonal features. Learning curves of \gpts{} on BLIMP challenges, obtained by spectral clustering and colored by fields.}
    % \vspace{-0.25cm}}
    \label{fig:pheno_clusters}
\end{figure*}
\section{Classifying the Learning Trajectories} \label{sec:types}

%We have investigated what and when do networks learn. 
To understand the latent features learned by NLMs, we categorize linguistic phenomena through the lens of their learning trajectories. We ask whether linguistically similar phenomena are learned in a similar fashion, and whether what is learned similarly is defined by linguistic terms.
% we take a closer look at the challenge's individual learning curves. 

%The $67$ BLIMP challenges are originally categorized into $13$ super-phenomena (e.g., island-related or quantifiers), which comprise $4$ fields, we analyze next. % (morphology, semantic, syntactic and syntactic-semantic)
We inspect linguistic categories by comparing the learning trajectories of their phenomena. In the Morphology field, we find that they display similar gradual curves, ultimately reaching high performance (median accuracy $0.85$, see Fig.~\ref{fig:field_morphology}). This may indicate that some latent features learned are morphological, and affect performance on almost all 'Morphology' phenomena.  

Syntax-semantics phenomena also present unique behavior: their scores plateau near chance performance (see Fig~\ref{fig:field_syntax_semantics}), suggesting that the learned features are insufficient to correctly represent phenomena in this field. The other fields, "semantics" and "syntax" (Figs~\ref{fig:field_semantics},\ref{fig:field_syntax}), do not present prototypical learning curves, suggesting that they are too broad to correspond to a single learning pattern. This, in turn, may suggest that they do not all correspond to a well-defined set of latent features.

Next, we follow the reverse direction and cluster the learning curves of \gpts{}. We use spectral clustering with 10 clusters and \href{https://scikit-learn.org}{sklearn} default parameters, by projecting the learning curves into a normalized Laplacian and applying k-means. Intuitively, learning curves with similar values along the principal directions, are clustered together. Other clustering methods show similar results.%\footnote{All hyperparameters follow the defaults in \href{https://scikit-learn.org/}{sklearn}.}%\footnote{\url{https://scikit-learn.org/}}

% The learning curves of some clusters are presented in Fig.~\ref{fig:pheno_clusters} (all the curves are given in App.~\S\ref{app_sec:all_clusters}). 
The clusters (Fig.~\ref{fig:pheno_clusters} and App.~\S\ref{app_sec:all_clusters}) reflect several learning profiles, some more expected than others. For some, accuracy improves as learning progresses (see Fig.~\ref{fig:cluster_start_random_and_improve}). Some are barely learned, and accuracy remains at near-chance level 
% during the entire course of learning
(see Fig.~\ref{fig:cluster_near_change_level}). Perhaps more surprisingly, some clusters deteriorate, and accuracy drops to nearly $0$ as learning progresses (see Fig.~\ref{fig:cluster_deteeriorate}). Notably, some challenges are quite easy -- NLMs instantly reach perfect accuracy (see Fig.~\ref{fig:cluster_perfect_accuracy}), while some are confusing -- NLMs performance is worse than chance (see Fig.~\ref{fig:cluster_deteeriorate}). In the latter cases, the NLMs presumably learn unrelated, harmful generalizations.

When inspecting the emerging clusters, many (but not all, see Fig.~\ref{fig:cluster_near_change_level}) contain a shared prominent field, but often varied super-phenomena (see Fig.~\ref{fig:cluster_start_random_and_improve}). %Hence, NLMs might be better characterized by (sub)fields than by super-phenomena.
% While the prominent field of this cluster is morphology combined with some semantic, it is composed of $6$ different super-phenomena (see App.~\S\ref{app_sec:all_clusters}).
Thus, while the categorization in BLIMP reflects a common linguistic organization of grammatical phenomena, from the perspective of learning trajectories only few of the super-phenomena in BLIMP show consistent behavior. %While humans see some challenges as requiring similar notions to understand (e.g., the NPI licensing challenges), it seems that NLMs see them as requiring distinct features. 
We cautiously conclude that there is some discrepancy between the common linguistic categorization of grammatical phenomena and the categorization induced by the learning trajectories of NLMs. An interesting direction for future work would therefore be the development of a theory that can account for the patterns presented by NLMs' learning trajectories.

%Looking for non-linguistic generalizations, we manually inspect a few phenomena with strong initial performance that then deteriorates. We find that some challenges are solvable by a simple rule, or by $n$-gram LMs. While $n$-grams cannot capture the grammatical phenomenon in general, they do solve the challenge. 

We manually inspect a few phenomena with strong initial performance that then deteriorates. We find that some of these challenges are solvable by a simple rule, easily learnable by an $n$-gram model.
For example, in "principle A case 1", always preferring subjective pronouns (e.g., "she" or "he") over reflexive ones (e.g., "himself", "herself") is sufficient to obtain a perfect score, and preferring "not ever" over "probably/fortunately ever" solves "sentential negation NPI licensor present". The fact that NLM performance deteriorates, fits our finding that nascent NLMs resemble an $n$-gram model. %, and that later potentially more structural generalizations are made. 
%In both examples, 5-gram and early-stage \gpts{} outperform the final score of GPT2$_{large}$. 

%In other challenges, it was difficult for us to explain what simple heuristic may solve the challenge. See for example the "inanimate" challenge in Table \ref{tab:BLIMP}.

%%%%%%%%%%%%%%%%%%%%%%%%%%%%%%%%%%%%%%%%%%%%%%%%%%%%%%%%%%%%%%%%%%
\section{Related Work}\label{sec:related_work}

Characterizing what networks learn is a long-standing challenge. Recently, studies suggested methods to analyze trained models such as probing \citep{tenney2018what, slobodkin2021mediators}, analyzing attention heads \citep{voita2019analyzing, abnar-zuidema-2020-quantifying} and neurons \citep[finding also correlations across epochs;][]{bau2018identifying} and assessing the extent to which LMs represent syntax \citep{van2019quantity}. Other works compare outputs, like us, to assess network generalizations \citep{choshen-abend-2019-automatically, Ontanon2021MakingTS}, look for systematic biases \citep{choshen-abend-2018-inherent,stanovsky-etal-2019-evaluating} or evaluate characteristics of outputs \citep{Gehrmann2021TheGB, Choshen2020ClassifyingSE}. \citet{McCoy2020BERTsOA} fine-tuned BERT and tested generalizations on the adversarial dataset HANS \citep{mccoy2019right}, finding models to make inconsistent generalizations. Their results differ from ours, but so is their setup, which involves fine-tuning for inference.

Characterizing the features learned by networks according to the order in which examples and phenomena are learned is a relatively new topic.
%in the machine learning literature
Recently, \citet{hacohen2019all, hacohen2021principal, Pliushch2021WhenDC} showed that classifiers learn to label examples in the same order. While their focus was on computer vision, it provided motivation for this work.
Other studies use learning dynamics as a tool, rather than a topic of study. They choose training examples \citep{toneva2018empirical}, categorize examples \citep{swayamdipta2020dataset} or characterize the loss-space \citep{xing2018walk}. Little research on NLM learning dynamics and generalization types was previously conducted. 

Perhaps the closest to this work is \citet{saphra2019understanding}, which compared LSTM representations with 3 types of linguistic tagger outputs, finding that correlation is low and that later in training, more context is used. The latter is reminiscent of our findings in \S\ref{sec:phases}.

In parallel work, \citet{Liu2021ProbingAT} probe models during training. They show that, early in training, information required for linguistic classifications is found somewhere in the layers of the model. Our work supports their findings by showing that grammar learning experiments conducted with one model are likely to replicate on another. Our methodology differs from theirs in requiring the information the model learnt to manifest itself in behavior rather than to be extractable with a dedicated classifier.

Studying the trajectories of language learning is a mostly untapped area in NLP, but is a long-established field of research in linguistics and psychology. Such lines of research study topics such as acquisition of phonemes \citep{kuhl1992linguistic}, morphology \citep{Marcus1992OverregularizationIL}, complex constructions \citep{Gropen1991AffectednessAD,Qingmei2007TheCA} and innate learning abilities \citep{Tomasello2003ConstructingAL}. Considerable computational work was also done on constructing models that present similar learning trajectories to those of infants \citep[among many others]{mcclelland1981interactive,perfors_tenenbaum_wonnacott_2010,Abend2017BootstrappingLA}.

Our work suggests that the generalizations NLMs make are coupled with the bottom-line performance. This gives a new angle and opens avenues of research when combined with previous work about bottom-line performance.
For example, the bottom-line performance of small models could predict the performance of larger models \citep{Ivgi2022ScalingLU}. In such cases, the type of generalizations made might also be predicted from the smaller models.

Our work is also closely related to fields such as curriculum learning \citep{bengio2009curriculum,hacohen2019power}, self-paced learning \citep{kumar2010self,tullis2011effectiveness}, hard data mining \citep{fu2017easy}, and active learning \citep{krogh1994neural,hacohen2022active, ein-dor-etal-2020-active}. In these fields, the order in which data should be presented to the learner is investigated. On the other hand, in our work, we study the order of the data in which the learner is learning -- which may shed some light on the advancement of such fields.

%%%%%%%%%%%%%%%%%%%%%%%%%%%%%%%%%%%%%%%%%%%%%%%%%%%%%%%
\section{Summary and Conclusions}\label{sec:conclusion}

We showed that NLMs learn English grammatical phenomena in a consistent order, and subsequently investigated the emerging trajectory. Our findings suggest that NLMs present consistent and informative trends. This finding suggests a path for studying NLMs' acquired behavior through their learning dynamics, as a useful complementary perspective to the study of final representations.  

Future work will consider the impact of additional factors, architectures and learning phases that appear only later in training. 
We hope that this work will increase the affinity between the knowledge and methodologies employed in developmental studies, and those used for studying NLMs. Our goal is to obtain a better understanding of what makes linguistic generalization complex or simple to learn, for both humans and NLMs.

%Moreover, even in the released, open questions are in abundance. 
%What is actually learned when phenomena decrease, what makes phenomena similar in the eye of the model and how are all those in comparison to language acquisition processes.
\FloatBarrier

\section*{Acknowledgments}
We thank Prof. Inbal Arnon for her helpful discussions.
This work was supported in part by the Israel Science Foundation (grant no. 2424/21), by a grant from the Israeli Ministry of Science and Technology, and by the Gatsby Charitable Foundations.

\clearpage
\bibliography{acl2020}
\bibliographystyle{acl_natbib}

\clearpage
\appendix
\section{Per challenge Graphs}\label{per_challenge}
We include behaviours of each model trained over the main dataset used (Wikipedia and books) on each BLIMP challenge by perplexity. In general, accuracy is similar despite different initialization and size of the GPT2 models. TransformerXL shows a similar trend, despite the uncomparable Perplexity. We supply several examples here and leave the rest to the data accompanying this paper.
\begin{figure}[htbp]
    \includegraphics[width=7.5cm]{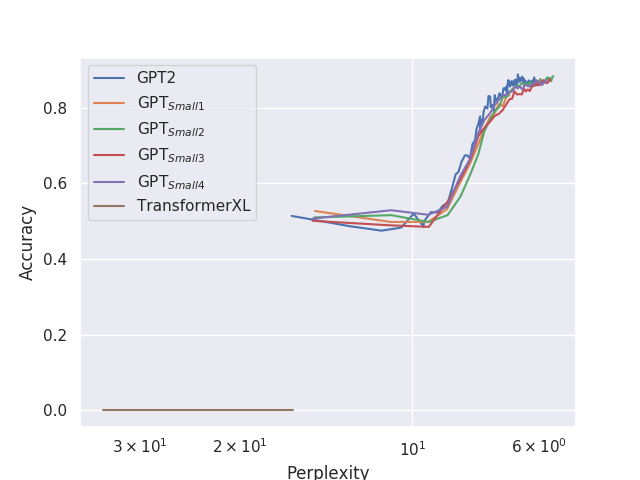}
    \caption{The accuracy on determiner noun agreement during training. Accuracy is similar despite different initialization and size of the GPT2 models. TransformerXL shows a similar trend, despite the uncomparable Perplexity.}
    \label{apfig:challenge_example}
\end{figure}

\begin{figure}[htbp]
    \includegraphics[width=7.5cm]{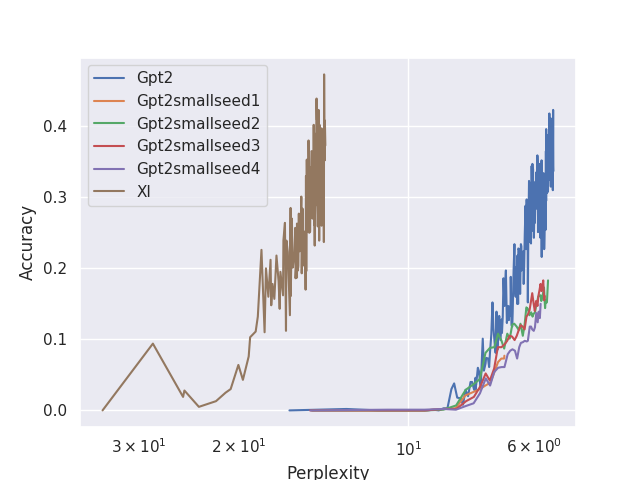}
    \caption{The accuracy on wh vs that with gap during training. \label{fig:wh_vs_that_with_gap_steps}}
\end{figure}

\begin{figure}[htbp]
    \includegraphics[width=7.5cm]{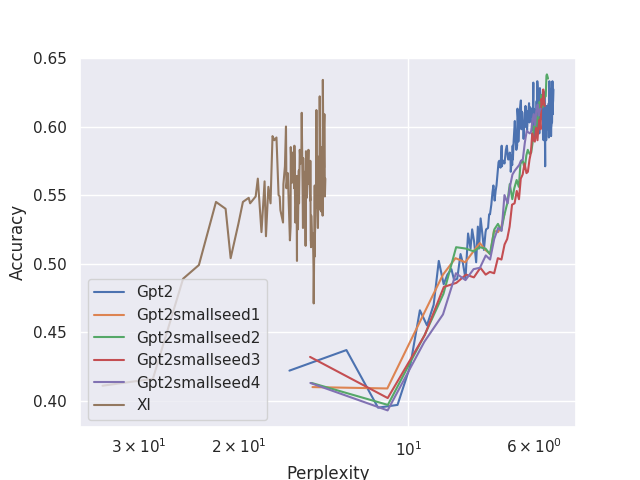}
    \caption{The accuracy on causative during training.}
\end{figure}

\section{Details on experimental settings}\label{sec:settings}
We include further settings to ensure reproduciblity of the results.
Parameters shared by all the trained NLMs include $32K$ tokens in the vocabulary, $5\cdot 10^{-5}$ learning rate, max gradient norm of $1$, Adam optimizer \citep{Kingma2015AdamAM}, and $10K$ warm-up steps. TransformerXl vocabulary is kept to its default. All other parameters, including \gpt{} size parameters, are the defaults according to the \href{https://huggingface.co/transformers}{HuggingFace} transformers library.%\footnote{\url{https://huggingface.co/transformers}}

Our 2-5 grams are KenLM \citep{Heafield2011KenLMFA} trained on WikiBooks. A second 5-gram model trained on GigaWord corpus \citep{graff2003english}, as reported by BLIMP. The Uni-gram LM is defined according to the frequency of a word in WikiBooks. Sentence probability is normalized by the number of words, which is helpful for the rare cases where the minimal pairs are of different lengths.

\section{Correlation during training}\label{sec:during_train}
\begin{figure}[htbp]
    \centering\includegraphics[width=0.8\columnwidth]{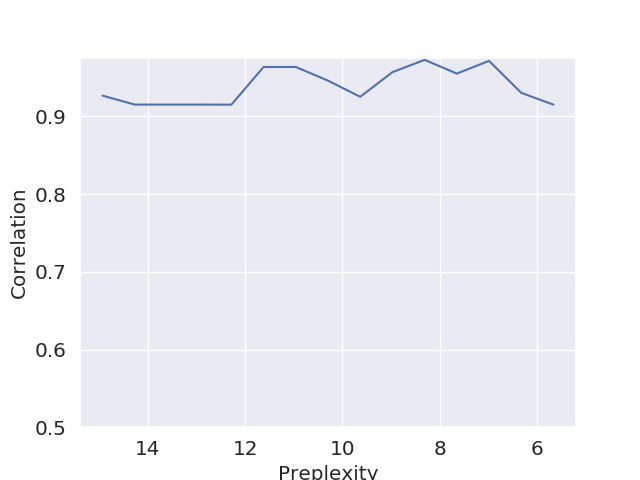}
    % \captionsetup{aboveskip=0pt, belowskip=0pt}
    \caption{Correlation between the performance vectors of \gpts{} and \gpt{}, aligned by perplexity.}
    \label{fig:correlation_by_step_with_base}
\end{figure}

We see that tendencies during training are not only similar between instances of the same architecture but also between different architectures. 
On comparable stages of learning, the \gpts{} and \gpt{} correlate well (>0.9) with respect to their performance vectors. We present the correlations of \gpts{} compared to \gpt{} in Fig.~\ref{fig:correlation_by_step_with_base}. We find the two learn in a similar order throughout their training.

We manually compare the results to TransformerXL. Qualitatively, observing the trajectories per challenge (Trajectories are found in Supp. \S\ref{per_challenge} and the supplied data) of TransformerXL, it seems to share the general tendencies of the GPT2 architectures. However, reaching a lower stage of training, it never improves on some challenges (e.g., determiner-noun agreement).

\section{Models are consistent on per example level}\label{sec:example_corr}

We compute the binary score of every example by each model. We reframe the question as an annotator agreement problem and ask whether the models agree on the right answer for each example. Framed this way, the methodology is clear. We compute Fleiss kappa \citep{Fleiss1969LargeSS} and find the per example correlation. The full results per step and challenge are added as a supplemental file. The average overall kappa is 0.83, models not only agree on the order of learning phenomena but also on the order of learning examples within each per-phenomenon (if learnt at all). While there are phenomena with lower and higher agreement, there are only two phenomena in the range of 0.5-0.6 agreement. Meaning even the most different ones have high example correlation and there is little variance between models to explain. 

Our main aim in this work is to compare models acquisition. However, we see the per example order of acquisition as less informative, unless we can cluster or name the examples learnt. The reason to choose the phenomena was to extract such names, and we hence focus in our work on them. 

Note, that consistency per example was shown before in the scope of computer vision \citep{hacohen2019all}. However, a critical difference is that they deal with classification and check whether which examples are learnt first. We however, aim to ask about generalization, given that you learn one task (language modelling), what type of generalizations do you make, tested on another. For example, while learning to predict the next word, the network understands after X steps that the verb should be in agreement with the subject. 

\section{Reproducing with other models}\label{sec:other_models}
\begin{figure}[tbhp]
    \includegraphics[width=7.5cm]{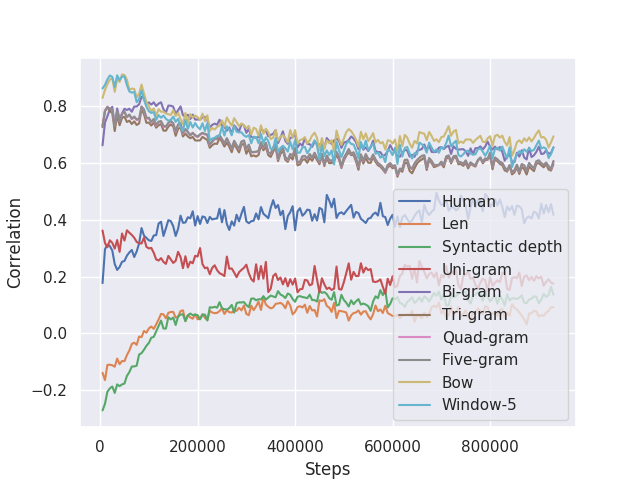}
    % \captionsetup{aboveskip=0pt, belowskip=0pt}
    \caption{Correlation between the difficulty of GPT2 and of other models for each phenomena in each training step.}
    \label{fig:bigmodel_metrics}
\end{figure}

\begin{figure}[tbhp]
    \includegraphics[width=7.5cm]{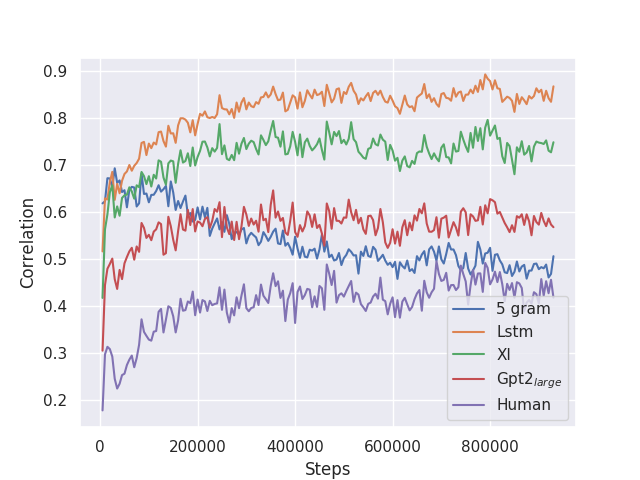}
    % \captionsetup{aboveskip=0pt, belowskip=0pt}
    \caption{Correlation between the difficulty predicted by BLIMP models and the difficulties for the model for each phenomena in each training step.}
    \label{fig:big-blimp-metrics}
\end{figure}

\begin{figure}[tbhp]
    \includegraphics[width=7.5cm]{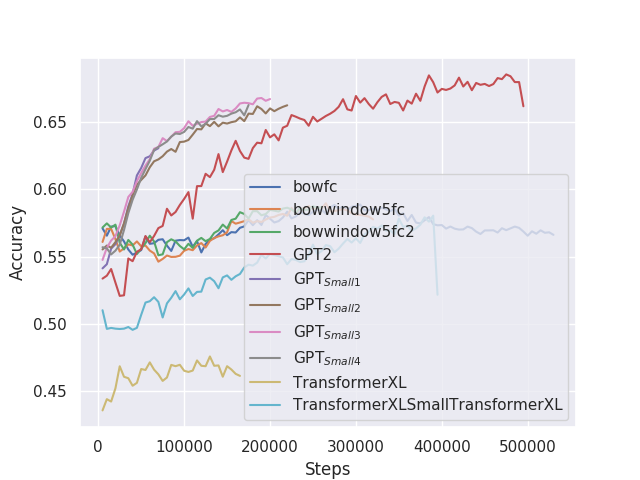}
    % \captionsetup{aboveskip=0pt, belowskip=0pt}
    \caption{Overall BLIMP accuracy by step.}
    \label{fig:accuracies}
\end{figure}

We provide the \gpt{} correlation with other models and with various metrics and models in Fig. \ref{fig:bigmodel_metrics} and \ref{fig:big-blimp-metrics} respectively. We also supply the average BLIMP accuracies of the models we trained in Fig. \ref{fig:accuracies}.

\subsection{Results mainly replicate in TransformerXL}\label{sec:xl}
We replicate the same experiment over the training of the TransformerXL instance. The TransformerXL seems to reach a lower stage of learning, probably due to the vast vocabulary and model. 

The model replicates some of the general notions seen on \gpt{}.
It correlates most with simpler models, then with humans and then with global features. At first, sentence length makes a sentence more challenging than its actual structure, 5 window BOW starts as more relevant than BOW over all the sentence.

We do see that the overall graph is quite straight. With that, the increase in correlation with humans is quite small, the BOW models don't drop and the evidence of relying on more abstract knowledge in late stages is less apparent. This might be expected, as we know the network reached an early step on the performance scale.

\begin{figure}[tbhp]
    \includegraphics[width=7.5cm]{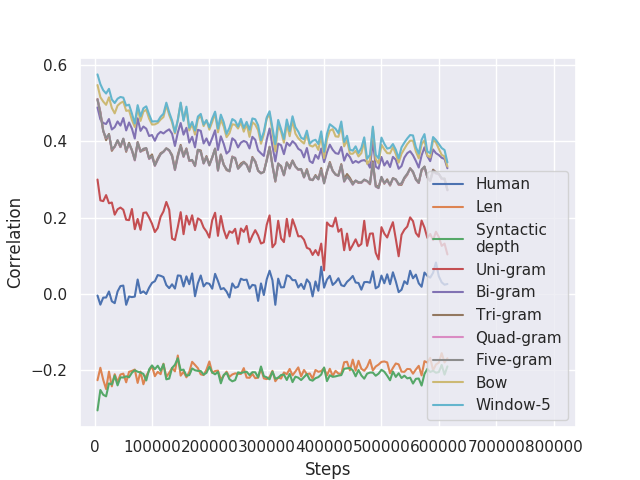}
    % \captionsetup{aboveskip=0pt, belowskip=0pt}
    \caption{Correlation between the difficulty predicted by metrics and the difficulties for the model for each phenomena in each time step.}
    \label{fig:xlmetrics}
\end{figure}

\section{Reproducing with other data}\label{sec:other_data}
\begin{figure}[tbhp]
    \includegraphics[width=7.5cm]{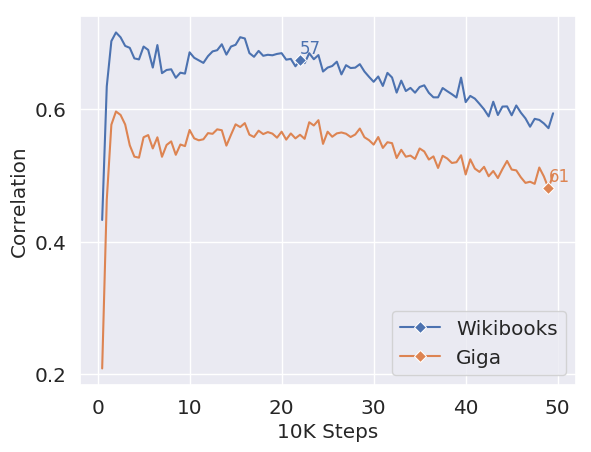}
    \caption{Correlation during training of \gpts{} on OpenWebText compared to 5-gram model trained on WikiBooks and on GigaWord. Correlation is over BLIMP challenges. Numbers indicate the overall average of the reference models over BLIMP and are found over the step with most similar accuracy on \gpts{}. \gpts{} best score is 67.}
    \label{fig:web5gram}
\end{figure}

As comparison to the correlations with our main model, we provide the correlations of \gpts{} trained on OpenWebText with the two 5-gram models, one on WikiBooks and one on Giga word (Fig. \ref{fig:web5gram}). We see that the higher resemblence to WikiBook trained model is kept despite being trained on the same data, but the difference is lower at the beginning and more stable. It might be the case that over reliance on the specific data is shown at those first steps where the difference is large, but it would require further evidence.

\begin{figure}[tbhp]
    \includegraphics[width=7.5cm]{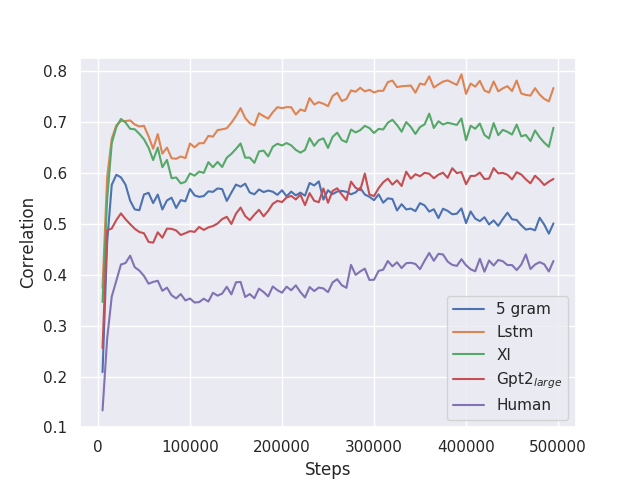}
    \caption{Correlation during training of \gpts{} trained on OpenWebText data compared to Off-the-shelve models and XL smaller models. The correlation with itself during training is shown in gray. Correlation is over BLIMP challenges. Numbers indicate the overall average of the reference models over BLIMP and are found over the step with most similar accuracy on \gpts{}. \label{fig:webtxt_cor}}
\end{figure}

We also compare the model to several other trained models in Fig. \ref{fig:webtxt_cor}.

\section{5-grams notes}\label{sec:5grams}
The gap between the correlation with the two 5-grams decreases during the first 50K steps or so, and then remains constant. This suggests that the choice of a dataset is more important during early NLM training.
Because, at the beginning the network learn generalizations which are more common to counts of one (huge, general domain) dataset than another, and this effect diminishes.
Possibly, this is because at this point NLMs rely more on word identity, rather than on abstract generalizations, that are shared to a greater extent across corpora (see \S\ref{sec:phases}).
We observe that the 5-gram trained on WikiBooks correlates better with \gpts{}, even when \gpts{} is not trained on it (not reported). % than the 5-gram trained on GigaWord, regardless of the NLM's training data. 
We cannot offer a simple explanation for this trend.

\section{Clustering BLIMP}\label{app_sec:all_clusters}
\begin{figure}[htbp]
    \includegraphics[width=7.5cm]{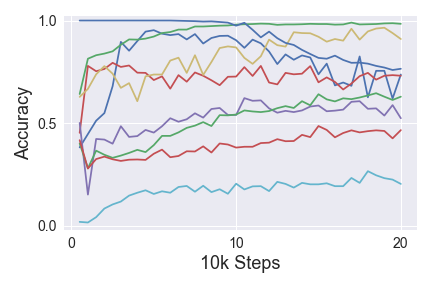}
    \caption{Cluster of semantic phenomena, each line is the trajectory of learning of a phenomenon.}
    \label{app_fig:cluster_semantics}
\end{figure}

\begin{figure}[htbp]
    \includegraphics[width=7.5cm]{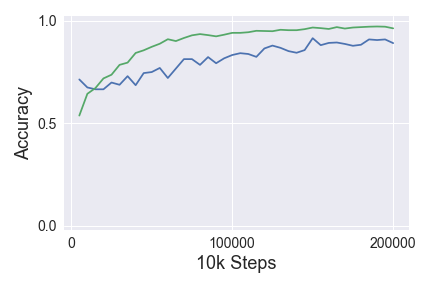}
    \caption{Anaphor agreement super phenomena trajectories. }
    \label{app_fig:cluster_anaphor_agreement}
\end{figure}

\begin{figure*}[bhtp]
    \begin{subfigure}{.46\textwidth}
      \centering
      \includegraphics[width=1\linewidth]{spectral_clusters/clusters/field0.png}
    \end{subfigure}
    \begin{subfigure}{.46\textwidth}
      \centering
      \includegraphics[width=1\linewidth]{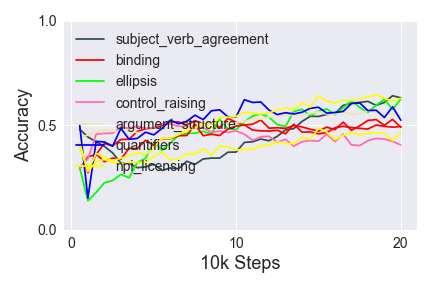}
    \end{subfigure}
    \caption{Cluster of phenomebna chosen by spectral clustering. The phenomena behave similarly but do not follow the same linguistic categorizations.}
    \label{app_fig:cluster0}
\end{figure*}

We include the learning curves of \gpts{} on BLIMP dataset, clustered according to fields (Fig.~\ref{app_fig:cluster_semantics}),%\ref{app_fig:cluster_morphology},\ref{app_fig:cluster_syntax},\ref{app_fig:cluster_syntax_semantics})
, super-phenomena (Fig.~\ref{app_fig:cluster_anaphor_agreement})%,\ref{app_fig:cluster_argument_structure},\ref{app_fig:cluster_binding},\ref{app_fig:cluster_control_raising},\ref{app_fig:cluster_determiner_noun_agreement},\ref{app_fig:cluster_ellipsis},\ref{app_fig:cluster_filler_gap_dependency},\ref{app_fig:cluster_irregular_forms},\ref{app_fig:cluster_island_effects},\ref{app_fig:cluster_npi_licensing},\ref{app_fig:cluster_quantifiers},\ref{app_fig:cluster_s-selection},\ref{app_fig:cluster_subject_verb_agreement})
, and the spectral clustering (Fig.~\ref{app_fig:cluster0}). %,\ref{app_fig:cluster1},\ref{app_fig:cluster2},\ref{app_fig:cluster3},\ref{app_fig:cluster4},\ref{app_fig:cluster5},\ref{app_fig:cluster6},\ref{app_fig:cluster7},\ref{app_fig:cluster8},\ref{app_fig:cluster9}) presented in \S5. The rest of the graphs are found in the data attached to this paper.
Due to restrictions on appendix files the figures are found in corresponding folders in the supplied data.

% \begin{figure}[htbp]
%     \includegraphics[width=7.5cm]{graphs/spectral_clusters/field/morphology.png}
%     \caption{}
%     \label{app_fig:cluster_morphology}
% \end{figure}

% \begin{figure}[htbp]
%     \includegraphics[width=7.5cm]{graphs/spectral_clusters/field/syntax.png}
%     \caption{}
%     \label{app_fig:cluster_syntax}
% \end{figure}

% \begin{figure}[htbp]
%     \includegraphics[width=7.5cm]{graphs/spectral_clusters/field/syntax_semantics.png}
%     \caption{}
%     \label{app_fig:cluster_syntax_semantics}
% \end{figure}

\FloatBarrier

\end{document}